\documentclass[twoside]{article}
\usepackage[accepted]{template}

\usepackage{url}
\usepackage{graphicx} 
\usepackage{subfigure}
\usepackage{amsmath}
\usepackage{amsfonts}
\usepackage{epstopdf}
\usepackage{algorithm}
\usepackage{algorithmic}
\usepackage{bbm}

\newtheorem{theorem}{Theorem}
\newtheorem{mydef}{Definition}

\begin{document}

\twocolumn[

\aistatstitle{Graph Kernels via Functional Embedding}

\aistatsauthor{ Anshumali Shrivastava \And Ping Li }

\aistatsaddress{Department of Computer Science \\ Computing and Information Science \\ Cornell University \\ Ithaca, NY 14853, USA\\ \texttt{Email:anshu@cs.cornell.edu}\And Department of Statistics and Biostatistics\\ Department of Computer Science \\ Rutgers University \\ Piscataway, NJ 08854, USA\\ \texttt{Email:pingli@stat.rutgers.edu} } ]

\begin{abstract}
We propose a representation of graph as a functional object derived from the power iteration of the underlying adjacency matrix.  The proposed functional representation is a graph invariant, i.e., the functional remains unchanged under any reordering of the vertices. This property eliminates the difficulty of handling exponentially many isomorphic forms. Bhattacharyya kernel constructed between these functionals significantly outperforms the state-of-the-art graph kernels on 3 out of the 4 standard benchmark graph classification datasets, demonstrating the superiority of our approach. The proposed methodology is simple and runs in time linear in the number of edges, which makes our kernel more efficient and scalable compared to many widely adopted graph kernels with running time cubic in the number of vertices.
\end{abstract}

\section{Introduction}

Graphs are becoming ubiquitous in modern applications spanning bioinformatics, social networks, search, computer vision, natural language processing, etc. Computing meaningful similarity measure between graphs is a crucial prerequisite for a variety of  learning algorithms operating on graph data. This notion of similarity typically varies with the application. In designing similarities (e.g., kernels) between graphs, it is  desirable to have a measure which incorporates the rich structural information and is not affected by spurious transformations like reordering of vertices.

 Note that, in certain applications, graphs can come with additional label information such as node or edge labels~\cite{Mahe04,Costa10}. These additional annotations are not always available in every domain (e.g. social networks) and  are typically expensive to obtain. In this paper, we focus only on the basic graph structures, without assuming any additional information.

A common approach for computing kernels is to extract an explicit feature map from the graph, and then compute the kernel values via certain standard operations between  features (e.g., inner products).  This line of techniques typically make use of \textbf{graph invariants} ~\cite{Shawe-taylor93}  such as  eigenvalues of Graph Laplacian as features.  For example, ~\cite{Kondor08} which uses harmonic analysis techniques to extract a set of graph invariants. It was shown that a simple linear kernel, i.e., dot product between these graph invariant numbers,   outperforms many other graph kernels. 

Alternatively, one can  design a kernel function $K(G_1,G_2)$  given graphs $G_1$ and $G_2$, directly using ``similarity'' between them~\cite{Vishwanathan06,Vishwanathanjlmr10}. For example, the random walk kernel~\cite{Gartner03,Kashima03}  is based on counting common random walks between two given graphs.
Another example is the shortest-path kernel~\cite{Borgwardt05} which is based on counting pairs of vertices, between two graphs, having similar shortest distance between them.

Although random walk kernels and path based kernels are still among the widely adopted  graph kernels, one common disadvantage with them is that walks and paths do not capture information of the substructures present in the graph~\cite{Shervashidze09,Kriege12}. To address this problem, a flurry of interest arose on kernels based on counting common subgraph patterns. Counting all possible common subgraphs was known to be $\mathbb{NP}$-complete~\cite{Gartner03}. This led to the development of graph kernels focusing only on counting small subgraphs; for example, ~\cite{Shervashidze09} counts common subgraphs with only 1, 2,  or 3 nodes also called as graphlets. This kind of technique is very popular in social network classification. Recently, ~\cite{UganderBK13} used histograms of size four subgraphs for classifying Facebook social networks.  However, simply counting common substructures like walks, paths, subgraphs, etc., ignores some crucial relative information between the substructures. For instance, the information of how different triangles are relatively embedded in the graph structure cannot be captured by simply counting the number of triangles. This relative information, as we show in this paper, is necessary for discriminating between different graph structures.

This paper follows an altogether different approach. We represent a graph as an expressive functional object. We first use the dynamical properties of the graph adjacency matrix to construct an informative summary of the graph. We then impose a probability distribution over the summary,
and we show that this distribution is a graph invariant. Bhattacharyya kernel between the obtained distribution, which we call {\bf Power Kernel}, significantly outperforms other well-known graph kernels on  standard benchmark graph classification datasets. In addition, we show that, unlike other kernels, most of which require $O(n^3)$ time to compute (where $n$ is the number of nodes), our kernel can be computed in time linear in the number of edges (which is at most $O(n^2)$). This makes the proposed methodology significantly more practical for larger graphs.

\section{Notation}

Given a graph $G$ with $n$ nodes, we denote its adjacency matrix by $A\in\mathbb{R}^{n\times n}$. In this paper, entries of $A$ are binary (0/1), i.e., $A(i,j) = 1$ means there is an edge between node $i$ and node $j$. We interchangeably use terms nodes and vertices, and terms graph $G$ and adjacency matrix $A$.  The graph will always be assumed to be unlabeled, undirected and unweighted with default $n$ number of nodes, unless otherwise specified. We use $\mathbbm{1}$ for a vector of all ones. By vector, we mean column vector, i.e., $n \times 1$ matrix.

To avoid overloading subscripts, we will follow  {\em Matlab} style notation while denoting rows and columns of a matrix. For a given matrix $A$, $A(i,:)$ will denote the $i^{th}$ row of $A$, while $A(:,i)$ will refer to the $i^{th}$ column. For a vector $x$, $x(i)$ will denote its $i^{th}$ component.

Every permutation $\pi:\{1,2,..,.n\} \rightarrow \{1,2,..,.n\}$ is associated with a corresponding permutation matrix $P$. One important property of a permutation matrix is that its transpose is equal to its inverse, $P^T = P^{-1}$. The effect of left multiplying a given matrix $A$ by $P$ shuffles its rows according to $\pi$, i.e., the $\pi(i)^{th}$ row of $PA$ is  $i^{th}$ row of $A$.  The effect of right multiplying has the same effect on columns instead of rows. For any permutation matrix $P$, graphs represented by adjacency matrices $A$ and $PAP^T$ are isomorphic, i.e., they represent the same graph structure except that the nodes are reordered according to $\pi$.



\section{Graphs as ARMA Models and Random Walk Kernels}

One way of representing graphs is to think of adjacency matrix $A \in \mathbb{R}^{n \times n}$ as a matrix operator operating in $\mathbb{R}^n$. A natural way of characterizing an operator is to see how it transforms a given vector $v \in \mathbb{R}^n$.  This idea was pioneered in case of graphs by works on diffusion kernels~\cite{Kondor02} followed by Binet-Cauchy kernels~\cite{vishwanathanSV07}. Here, the adjacency matrix was treated as a dynamical system and similarity measure between these systems was used as a similarity between corresponding graphs.

%

 In~\cite{vishwanathanSV07}, graph with adjacency matrix $A$ was associated with the following noiseless ARMA model.
\begin{equation}
\label{eq:ARMA}
\begin{array}{l}
\displaystyle y_t =x_t ;  \hspace{10 mm} x_{t+1} = Ax_t.
\end{array}
\end{equation}
It was shown that the random walk kernel between two graphs, with adjacency matrix $A$ and $A'$, is actually the Binet-Cauchy trace kernel over the corresponding ARMA models that takes the following form (see Eq. (10) in~\cite{vishwanathanSV07}):
\begin{equation}
K(A,A') = \sum_{t=1}^\infty e^{-\lambda t} y_t^TWy_t'.
\label{eq:ARMAKernel}
\end{equation}
where  $x_0 = \mathbbm{1}/ |V|$, $x_0' = \mathbbm{1}/|V'|$ and $W$ is a matrix of all ones. The discounting term $ e^{-\lambda t}$ is necessary for the finiteness of the summation. Fortunately, the infinite summation in Eq. (\ref{eq:ARMAKernel}) has a closed form solution and can be computed in $O(n^3)$~\cite{Vishwanathan06}.

 It can be observed from Eq. (\ref{eq:ARMAKernel}) that random walk kernel  is simply a discounted summation of similarity between $y_t$ and $y_t'$, where the summation is taken over $t$.  It does not take into account the covariance structure of the dynamical system.  In particular, given the adjacency matrix A, if we think of $\{y_t: t \in \mathbb{N}\}$ as a series, one of the identifying characteristics of a series is how $y_t$ relates with $y_{t'}$ for $t \ne t'$. Such kind of auto-covariance structures are very crucial in time series modeling literature. Unfortunately, this information is not taken into consideration while computing the similarity in Eq. (\ref{eq:ARMAKernel}).

There are more expressive kernels for ARMA models like the determinant  kernel~\cite{vishwanathanSV07}. However, determinant kernel for ARMA models are not applicable for graphs because it is sensitive to reordering of rows~\cite{wolf03}.  It should be noted that given a permutation matrix $P$ and an adjacency matrix $A$, $A$  and $PAP^T$ leads to different dynamical systems but the graphs represented by them are isomorphic.  Therefore, we need a very different approach for defining kernels between graphs which takes into account the covariance structure of the series $\{y_t: t \in \mathbb{N}\}$.

We proceed by computing an isomorphic invariant functional  representation of a given graph, which captures the covariance information of the dynamical system. We describe this functional embedding in the next section.



%

\section{Graph Embedding in Functional Space}


In Eq. (\ref{eq:ARMA}),  $y_t$ is simply a power iteration of matrix $A$.  A small history of power iteration often captures sufficient information about the underlying matrix~\cite{lin2010power}. Our representation capitalizes on this fact. We first extract a summary of power iteration as shown in Algorithm~\ref{alg:summary}. In standard power iteration, we start with a given normalized vector $x^{(0)}$ and at each iteration $t \in \{1,2,...,k\}$, we generate vector $x^t =  A \times \frac {x^{(t-1)}}{||x^{(t-1)}||_1}$ recursively.  The choice of normalization is not important.

\begin{algorithm}[tb]
   \caption{\emph{Power Summary of Graph}}
   \label{alg:summary}
\begin{algorithmic}
   \STATE {\bfseries Input:} Adjacency matrix $A\in\mathbb{R}^{n \times n}$; initial vector $x^{(0)}\in\mathbb{R}^{n\times1}$; number of power iterations $k$

 \FOR{$t=1$ {\bfseries to} $k$}
   \STATE \hspace{0.3in}$x^{(t)} =   A \times \frac {x^{(t-1)}}{||x^{(t-1)}||_1}$

   \vspace{0.05in}

   \STATE \hspace{0.25in} $\mathbb{S}_{x^{(0)}}^{A}(:,t) =  x^{(t)}$
   \ENDFOR
  \RETURN  $\mathbb{S}_{x^{(0)}}^{A}$
\end{algorithmic}
\end{algorithm}

We refer the $(n \times k)$ matrix,  whose $j^{th}$ column corresponds to the $x^{j}$, as $\mathbb{S}_{x^{(0)}}^{A}$. $\mathbb{S}_{x^{(0)}}^{A}$ is not  permutation invariant because  $A^tx^{(0)}$ and $(PAP^T)^tx^{(0)}$, for general $x^{(0)}$,  are not equal. However, if the starting vector is $x^{(0)} = \mathbbm{1}$ (where $e$ is a vector of all ones), then reordering the nodes by permutation matrix $P$  just shuffles the rows of $\mathbb{S}_{\mathbbm{1}}^{A}$ in the same order. This fact can be stated as the following theorem.
\begin{theorem}

\label{thoe:1}
If P is any permutation matrix, then $\mathbb{S}_{\mathbbm{1}}^{PAP^T} = P \times\mathbb{S}_{\mathbbm{1}}^{A} $, and the all-one vector $\mathbbm{1}$ is the unique starting vector, up to scaling, having such property for all $A$ and $P$.
\end{theorem}
${\bf Proof}$ Using the identity$ P^T = P^{-1}$, it is not difficult to show that for any permutation matrix P, $(PAP^T)^k = PA^kP^T$. This along with the fact $P^T\times \mathbbm{1} = \mathbbm{1}$, yields the required result. For uniqueness, let $x^{(0)}$ have two different components at $i$ and $j$, then $PAP^Tx^{(0)} \ne PAx^{(0)}$ in general. Equality here forces a constraint on $A,P$ and $x^{(0)}$. Since we have limited degrees of freedom for $x^{(0)}$ compared to choices of $A$ and $P$,  this will end in contradiction. $\hfill\square$\\

One more intuitive way to see why Theorem~\ref{thoe:1}  is true is to disregard normalization and imagine that at time step $t=0$, we associate every node in the graph with the starting number $1$. During every iteration of Algorithm~\ref{alg:summary}, which is multiplication by $A$, we update this number on every node with the sum of numbers on all its neighbors. A simple recursive argument tells us that the sequence of numbers generated on each node, under this process, is not going to change as long as the neighborhood structure is preserved. Unit vector $\mathbbm{1}$ is the only starting choice that does not distinguish between nodes. In fact, each row vector of $\mathbb{S}_{\mathbbm{1}}^{A}$ can be treated as a representation for the corresponding node in the graph. Such kind of updates are very informative and is the motivation behind many celebrated link analysis algorithms including Hyper-text Induced Topic Search (HITS) ~\cite{Kleinberg98}.


In light of Theorem~\ref{thoe:1}, we can associate a set of $n$ vectors, corresponding to rows of  $\mathbb{S}_{\mathbbm{1}}^{A}\in\mathbb{R}^{n \times k}$, with graph $G$ as a permutation invariant representation. Our proposal, therefore, is a mathematical quantity that describes this set of vectors as a representation for graph.

We have two choices: 1) we can either think of a subspace represented by these $n$ vectors, or 2) we can think of these $n$ vectors as samples from some probability distribution~\cite{kondor2003kernel}.  This choice depends on size of $n$ and $k$. In case where $n$ is large compared to $k$,  the subspace represented by $n$ vectors of dimension $k$ will almost always be the whole $k$ dimensional Euclidean vector space, and it will not be very informative. On the other hand, when $k$ is large compared to $n$, the subspace representation may be more informative compared to fitting a distribution. For example, if we decide to fit a Gaussian distribution over these vectors, when $k$ is more than $n$, the covariance matrix is not very informative.

Power iteration converges very quickly due to its geometric rate of convergence. We therefore need much smaller values of $k$ compared to $n$. Hence, we associate a probability function with the rows of $\mathbb{S}_{\mathbbm{1}}^{A}$. We can get a variety of permutation independent functional embeddings by different choices of this distribution functions. We use the most natural distribution function, the Gaussian, for two major reasons: the similarity computations are usually in closed form and it nicely captures the correlation structure of $\mathbb{S}_{\mathbbm{1}}^{A}$.  Since we will always use $x^{(0)} = \mathbbm{1}$, for notational convenience we will drop the subscript $\mathbbm{1}$.

\begin{mydef}
Given an undirected graph $G$ with adjacency matrix $A\in\mathbb{R}^{n \times n}$ and $\mathbb{S}^{A}$ computed from Algorithm~\ref{alg:summary} run for $k$ iterations. Let $\mu^A \in \mathbb{R}^k$  be the mean of column vectors  of  $\mathbb{S}^{A}$ and $\Sigma^A \in  \mathbb{R}^{k \times k}$ denote the covariance matrix of     $\mathbb{S}^{A}$:
 \begin{align}\notag
&\mu^A = \frac{1}{n}\sum_{i=1}^n\mathbb{S}^A(i,:),\notag \\
&\Sigma^A = \frac{1}{n} \sum_{i=1}^n(\mathbb{S}^A(i,:) - \mu^A)(\mathbb{S}^A(i,:) - \mu^A)^T \notag.
  \end{align}
  We define  ${\bf \Psi^A \in (\mathbb{R}^k \Rightarrow \mathbb{R^+})}$  of graph $G$ as a probability density function of multivariate Gaussian with mean $\mu^A$ and covariance $\Sigma^A$
$$ \Psi^A(x) = \frac{1}{\sqrt{{(2\pi)}^{k} |\Sigma^A |}} e^{-\frac{(x- \mu^A)^T{\Sigma^A}^{-1}(x- \mu^A) }{2}}.$$
\end{mydef}

Since, our representation is defined as a Gaussian density over the bag of vectors, it can also be interpreted as a Gaussian Process, see~\cite{kondor2003kernel}. This $\Psi$ representation has the desired property that it is invariant under reordering of nodes.
\begin{theorem}
For any permutation matrix P we have $\Psi^A = \Psi^{PAP^T}$.
\label{theo:2}
\end{theorem}
${\bf Proof}$ Using Theorem~\ref{thoe:1}, it is not difficult to see that $\mu^A = \mu^{PAP^T}$ and $\Sigma^A = \Sigma^{PAP^T}$.$\hfill\square$\\
\vspace{-0.1in}

Although Theorem~\ref{theo:2} captures graph isomorphism in one direction, this representation is not an \emph{if and only if} relationship, and we cannot hope for it as we would have then solved the \emph{Graph Isomorphism Problem}.
 Although the complexity of \emph{Graph Isomorphism Problem} is still a big open question, for most of the graphs in practice it is known to be easy and a small summary of power iteration is almost always enough to discriminate between non-isomorphic graphs. In fact, real wold graphs usually possess very distinct spectral behavior~\cite{farkas2001spectra}. We can therefore expect the $\Psi$ embedding to be an effective representation for graphs encountered in practice.

It might seem little uncomfortable to call it a distribution because the row vectors of $\mathbb{S}^{A}$ never change, and so there is nothing stochastic. It is better to think of this representation of graph as an object in a functional space $(\mathbb{R}^k \rightarrow \mathbb{R^+})$. The distribution analogy gives the motivation of a mathematical object for a set of vectors, and a simple intuition as to why Theorem~\ref{theo:2} is true given Theorem~\ref{thoe:1}.


\section{The Proposed Kernel}
\label{sec:propkernel}



We define the {\bf Power Kernel} between two graphs with adjacency matrices $A$ and $B$ as a Bhattacharyya Kernel~\cite{jebara10} between ${\Psi}^A(x)$ and ${\Psi}^B(x) $
\begin{equation}
K(A,B) = \int_{\Omega}  \sqrt{{\Psi}^A(x)}\sqrt{{\Psi}^B(x)}dx.
\label{eq:bhatta}
\end{equation}

Since${\Psi}^A(x)$ and ${\Psi}^B(x) $ are pdf of Gaussians, Eq. (\ref{eq:bhatta}) has closed form solution given by:

\begin{align}
\label{eq:kernel}
 K(A,B) &= |\Sigma^A|^{-\frac{1}{4}}|\Sigma|^{\frac{1}{4}}|\Sigma^B|^{-\frac{1}{4}}  \notag\\
                 &\times e^{(T1+T2+T3)}\\
T1 &= -\frac{1}{4}(\mu^A)^T(\Sigma^A)^{-1}(\mu^A)\notag\\
T2 &= -\frac{1}{4}(\mu^B)^T(\Sigma^B)^{-1}(\mu^B)\notag\\
T3&= \frac{1}{2}\mu^T\Sigma^{-1}\mu\notag\\
 \Sigma &= \frac{\Sigma^A +\Sigma^B}{2}\notag\\
 \mu &=  \frac{1}{2}(\Sigma^A)^{-1}\mu^A+\frac{1}{2}(\Sigma^B)^{-1}\mu^B\notag
\end{align}

While designing kernels for graph ensuring positive semi-definiteness is not  trivial and many previously proposed kernels do not satisfy this property~\cite{Vishwanathanjlmr10,vert2008optimal}. Since our kernel is a kernel over well studied mathematical representation we get this property for free, which is an immediate consequence of the result that Bhattacharyya kernels are positive semidefinite.

\begin{theorem}
Power Kernel is positive semidefinite.$\hfill\square$
\end{theorem}

Overall, we have a very simple procedure for computing kernel between two graphs with adjacency matrices $A\in\mathbb{R}^{n_a \times n_a}$ and $B\in\mathbb{R}^{n_b \times n_b}$. The procedure is summarized in Algorithm~\ref{alg:powerkernel}.

\begin{algorithm}[tb]
   \caption{\emph{Power Kernel}}
   \label{alg:powerkernel}
\begin{algorithmic}
   \STATE {\bfseries Input:} A ($n_a \times n_a$), B ($n_b \times n_b$), $k$

   \vspace{0.05in}

   \STATE 1) Compute $\mathbb{S}^A$ and $\mathbb{S}^B$ using Algorithm~\ref{alg:summary} for $k$ iterations.

   \vspace{0.05in}

   \STATE 2) $\mu^A = \frac{1}{n}\sum_{i=1}^n\mathbb{S}^A(i,:)$

   \vspace{0.05in}

    \STATE 3) $\mu^B = \frac{1}{n}\sum_{i=1}^n\mathbb{S}^B(i,:)$

    \vspace{0.05in}

   \STATE 4) $\Sigma^A = \frac{1}{n} \sum_{i=1}^n(\mathbb{S}^A(i,:) - \mu^A)(\mathbb{S}^A(i,:) - \mu^A)^T$

   \vspace{0.05in}

   \STATE 5) $\Sigma^B =  \frac{1}{n}\sum_{i=1}^n(\mathbb{S}^B(i,:) - \mu^B)(\mathbb{S}^B(i,:) - \mu^B)^T$

   \vspace{0.05in}

    \STATE 6) Compute K(A,B) using Eq. (\ref{eq:kernel})

    \vspace{0.05in}

  \RETURN  K(A,B)
\end{algorithmic}
\end{algorithm}

The value of $k$ determines the number of power iterations in Algorithm~\ref{alg:summary}. For adjacency matrix $A$, let $\lambda_1 \ge \lambda_2 \ge ... \lambda_n$ be the eigenvalues  and $v_1, v_2, ... v_n $ be the corresponding eigenvectors.  The $t^{th}$ iteration on vector $x^{(0)}$ will generate $A^tx^{(0)} = c_1\lambda_1^tv_1 + c_2\lambda_2^tv_2 + ... + c_n\lambda_n^tv_n$,  where $(c_1,c_2, ..., c_n)$ is the representation of $x^{(0)}$ in the basis of $v_i$'s, i.e., $x^{(0)} = c_1v_1 + c_2v_2 + ... + c_nv_n$. This gives
\begin{equation}
A^tx^{(0)}  = \lambda_1^t\left[ c_1v_1 +  \sum_{i =2}^n\bigg(\frac{\lambda_i}{\lambda_1}\bigg)^t c_iv_i \right].
\end{equation}

We can see that power iteration looses information about the $i^{th}$ eigenvalue and eigenvector at an exponential rate of $(\frac{\lambda_i}{\lambda_1})^t$. A matrix is uniquely characterized by the set of its eigenvalues and eigenvectors, and we need all of them to fully capture the information in the matrix. It should be noted here, that unlike other machine learning applications where small eigenvalues corresponds to noise, in our case the  information of the whole spectrum is needed.  We therefore need small values of $k$ like 4 or 5. Larger values of $k$ will cause the information of the larger eigenvalues to dominate the representation, and this will make the kernel values biased towards the dominant spectrum of $A$.

\section{Running Time}
\label{sec:comp}
 We now analyze the running time of each step in Algorithm~\ref{alg:powerkernel}. For simplicity, let $n = \max(n_a,n_b)$.  Step (1) requires  running Algorithm~\ref{alg:summary} on both the graphs, which consists of matrix vector multiplications for $k$ iterations. The complexity of Step (1) is thus $O(n^2 k)$. Steps (2) and (3) compute the mean of $n$ vectors, each with dimension $k$, both of which cost $O(nk)$. Steps (4) and (5) compute the sample covariance matrix whose complexity is  $O(nk^2)$ each. The final step requires evaluating Eq. (\ref{eq:kernel}) on the computed mean and covariance matrices, which requires $O(k^3)$ operations. Overall, the total time complexity for computing the kernel from scratch is $O(n^2k+ nk^2 +k^3)$.

The recommended value of $k$ is usually a small constant (e.g. 4 or 5) even for large graphs. Treating $k$ as a constant, the time complexity is  $O(n^2)$ in the worst case. In fact, due to the sparsity of the adjacency matrix $A$,  the actual time complexity is $O(E)$, where $E$ is the total number of edges (which is at most $O(n^2)$). In other words, our total running time is linear in the number of edges.  The current state-of-the-art kernels including skew spectrum of graph~\cite{Kondor08}, random walk kernel~\cite{Gartner03}, require $O(n^3)$ computations while shortest path kernel~\cite{Borgwardt05} is even costlier. The graphs that we encounter in most real-world applications are in general very sparse, i.e., $E\ll n^2$.   Moreover, when the number of edges is  on the order of the number of vertices (which is not unusual),  our algorithm is actually linear in $n$. This makes our proposed power kernels scalable even for web applications.

Note that, we can pre-compute the first five steps in Algorithms~\ref{alg:powerkernel}, independently for each graph. After this preprocessing, kernel computation only requires $O(k^3)$ per pair, which is a constant.

\section{Why Covariance Captures Relative Information ?}

\label{sec:why}
The work of~\cite{Kondor08} was based on extracting permutation invariant features from graph using an algebraic approach. Our representation leads to a new set of invariants. As a consequence of Theorem~\ref{theo:2},  $\mu^A$ and $\Sigma^A$ are graph invariants.

Define ${\bf \mathcal{N}^i_t}$ as the number of disjoint paths of length $t$, in the given graph $G$, having node $i$ as one of its end points. In computing ${\bf \mathcal{N}^i_t}$, we allow repetition of nodes. One simple observation is that the $i^{th}$ component of $A^t\mathbbm{1}$, i.e. $A^t\mathbbm{1}(i)$, is equal to $\mathcal{N}^i_t$.  This fact can be proven by a simple inductive argument, where the base case ${\bf \mathcal{N}^i_1}$ corresponds to the degree of the node $i$.

The $t^{th}$ component of $\mu^A$ is the mean of $\mathcal{N}^i_t$, i.e. $$\mu^A(t) = C_1\sum_{i=1}^n\mathcal{N}^i_t,$$ which is a trivial graph invariant because it is the total number of paths of length $t$, in the given graph, multiplied by a constant. The constant $C_1$ comes into the picture due to normalization.

Interesting set of invariants come from the matrix  $\Sigma^A$.  The $(t_1,t_2)^{th}$ element of $\Sigma^A$ can be written as
$$\Sigma^A(t_1,t_2) = C_2 \sum_{i =1}^n (N_{t_1}^i - \mu^A(t_1))\times (N_{t_2}^i -\mu^A(t_2)) $$
which the correlations among the number of paths of length $t_1$ and that of length $t_2$ having a common endpoint. When $t_1 = t_2 = t$, it can be interpreted as the variance in the number of paths of length $t$ having a common endpoint. In the hindsight, its not difficult to see that these aggregated statistics of paths of different lengths starting at a given node are graph invariants. We will see in the next Section that this information is very useful in discriminating various graph structures.

$\mu^A(t)$ captures the information about the mean statistic of different kind of paths present in the graph. $\Sigma^A$ captures the relative structure of nodes in each graph. The correlations between various kinds of paths relative to a node indicated its relative connectivity in the graph structure. This kind of relative correlation information were missing in random walk kernels and path based kernels, which only count common paths or walks of same length between two given graphs. Even kernels trying to count common small subgraphs do not capture this relative structural information sufficiently. $\mu^A(t)$ and  $\Sigma^A$  capture aggregate behavior of paths relative to different nodes, and they can also be treated as an informative summary of the given graph.

Gaussian density function is just one way of exploiting this correlation structure.  We can generate other functionals on the rows of $\mathbb{S}^A$. For example, we can generate an expressive functional by using kernel density estimation on these set of  $n$ vectors, and Theorem~\ref{thoe:1} guarantees that the obtained functional is a graph invariant. We believe that such invariants can provide deeper insight which could prove beneficial for many applications dealing with graphs.  The behaviors of these graph invariants raise many interesting theoretical questions which could be of independent interest. For instance we can ask, ``what will be the behavior of these invariants if the graph has low expansion?"

\section{Experiments}

\begin{table*}[t]
\caption{Prediction Accuracy in percentage for power kernel and the state-of-the-art graph kernels on four classification benchmark  datasets. The reported results are averaged over 10 repetitions of 10-fold cross-validation. Standard errors are indicated using parentheses.}
\label{tab:result}

\centering
\begin{center}

\begin{sc}
\begin{tabular}{|p{5.8cm}| c| c| c| c|}
\hline
{\bf Datasets} & MUTAG & ENZYMES & NCI1 &NCI109 \\
\hline
{\bf No of Instances/Classes} & 188/2 & 600/6 & 4110/2 & 4127/2 \\
\hline
{\bf Max number of nodes} & 28 & 126 & 111 & 111 \\
\hline
Power Kernel  (This Paper) & 83.22(0.47)&  {\bf 34.60(0.48)}& {\bf 70.73(0.10)}  & {\bf 70.15(0.12)}\\
\hline
Reduced-Skew-Spectrum~\cite{Kondor08} &{\bf 88.61(0.21)}& 25.83(0.34)& 62.72(0.05)& 62.62(0.03)\\
\hline
Graphlet-Count-Kernel~\cite{Shervashidze09}  & 81.7(0.67)& 23.94(0.4)&54.34(0.04) & 52.39(0.09)  \\
\hline
Random-Walk-Kernel~\cite{Gartner03} & 71.89(0.66)& 14.97(0.28)& 51.30(0.23)&53.11(0.11)\\
\hline
Shortest-Path-Kernel~\cite{Borgwardt05}   & 81.28(0.45)& 27.53(0.29)& 61.66(0.10)& 62.35(0.13)\\
\hline
\end{tabular}
\end{sc}

\end{center}

\end{table*}

We follow the evaluation procedure of~\cite{Kondor08,Kondor09}. We chose the same four benchmark graph classification datasets consisting of the graph structure of the chemical compounds: MUTAG, ENZYMES, NCI1 and NCI109, used in~\cite{Kondor08,Kondor09} for their diversity in terms of size and as well as tasks. In each of these dataset, each data point is a graph structure associated with a classification label. MUTAG~\cite{debnath91} is a dataset of 188 mutagenic  aromatic and hetroaromatic nitro compounds, labeled according to whether or not they have mutagenic  effect on Gram-negative bacterium $Salmonell$ $typhimurium$. The maximum number of nodes in this dataset is 28 with mean around 19, while the maximum number of edges is 33 and the mean is around 20. ENZYMES is a dataset of protein tertiary structure, which was used in~\cite{Borgwardt205}. It consists of 600 enzymes from the BRENDA enzymes database~\cite{Schomburg04}. This is a multi-class classification task, where each enzyme has the label as to which of the 6 EC top level class it belongs to. The maximum number of nodes in this dataset is 126 with average around 32.6,  while the maximum number of edges is 149 and the mean is around 62. The other two balanced datasets,  NCI1 and NCI109, classify compounds based on whether or not they are active in an anti-cancer screen~\cite{Wale06}. For both NCI1  and NCI109 the maximum number of nodes is 111 with mean around 30, and the maximum number of edges is 119 with mean around 32.

 Our focus will remain on evaluating the basic structure captured by our functional representation $\Psi^A$. We therefore focus our comparisons with methodologies not relying on node and edge label information. We repeat evaluation procedure followed in~\cite{Kondor08,Kondor09} with power kernel. The evaluations consists of running kernel SVM on the four datasets using different kernel. The standard evaluation procedure used is as follows. First split each dataset into 10 folds of identical size. Combine 9 of these folds and again split it into 10 parts, then use the first 9 parts to train the C-SVM~\cite{libsvm} and use the 10th part as validation set to find the best performing value of C from $\{10^{-7},10^{-6},...,10^7\}$. With this  choice of C, train the C-SVM on all the 9 folds (form initial 10 folds) and predict on the 10th fold acting as an independent evaluation set. The procedure is repeated 10 times with each fold acting as an independent test set once. For each dataset the whole procedure is then repeated 10 times randomizing over partitions. The mean classification accuracy and the standard errors are shown in Table~\ref{tab:result}.

Since the results are averaged over 10 runs with different partitions, the numbers are very stable. We borrowed the accuracy values of state-of-the-art unlabeled graph kernels:  random walk kernel~\cite{Gartner03}, shortest path kernel~\cite{Borgwardt05}, graphlet count kernel~\cite{Shervashidze09}, and reduced skew spectrum of graph from~\cite{Kondor08,Kondor09}, where parameters, if any, for these kernels were optimized for best performance.

As noted before, the value of $k$ should not be large. Though we have the choice to tune this value for different datasets independently, to keep things simple and allow easy replication of results, we report the results for a fixed value of $k =5$ on all the four datasets.

From the results, e can see that other than the MUTAG dataset, power kernel outperforms other kernels on the remaining 3 datasets.  On NCI1 and NCI109, which are larger datasets with larger graphs compared to MUTAG, we beat the previous best performing kernel, which is based on skew spectrum of graph, by a huge margin. On these two datasets, power kernel gives a classification accuracy of around 70\% while the  best performing baseline can only achieve around 62\%. In case of ENZYMES dataset, the shortest path kernel performs the best among other baseline kernels and achieves 27.53\% accuracy, while we can achieve around 34.6\%.  This significant improvement clearly establishes the expressiveness of our representation in capturing structure of graphs.

On MUTAG dataset the accuracy of 88.61\% is achieved by reduced skew spectrum kernel while power kernel gives 83.22\%. We believe that this is due to the fact that MUTAG consists of relatively much smaller graphs, and it seems that the few graph invariant features generated by reduced skew spectrum sufficiently capture the discriminative information in this dataset.  On datasets with larger graphs, such features are less expressive than our functional representation, and hence power kernel leads to much better results.  Also MUTAG  dataset contains only 188 data elements, and so the percentage difference is not significant as compared to larger dataset like NCI1 and NCI109.

 We always outperform graphlet count kernel, random walk kernel and shortest path kernel. This shows that our basic representation is much more expressive and superior. It is not surprising because we are capturing higher order correlation information, while kernels based on counting common paths or subgraphs of small size miss this relative information. Dissecting graphs into small subgraphs looses a lot of information.

As shown in Section~\ref{sec:comp} our algorithm runs in $O(E)$ and from the statistics of the dataset we can see that on an average the edges are of the order of vertices, and so the running time complexity of power kernel in this case is actually around $O(n)$, while all other competing methods except graphlet count kernel require at least $O(n^3)$. Therefore, we have a huge gain in performance. The running time complexity of graphlet kernel is competitive with our method but accuracy wise our method is much superior. The whole procedure for power kernel is simple and since we haven't tuned anything except $C$ for SVM all these numbers are easily reproducible.

\section{Discussion: Effect of Perturbations}
\label{sec:why}

The success of isomorphism capturing kernels is due to their ability of preserving near neighbors with high probability. Our proposed power kernel posses the following two properties:
\begin{enumerate}
\item If two graphs $A$ and $B$ are isomorphic then $K(A,B) = 1$, and if they are not, then likely $K(A,B) < 1$.
\item If two graphs $A$ and $B$ are small perturbed versions of each other then $K(A,B)$ should be close to 1, in particular it should be higher compared to two random graphs.
\end{enumerate}
Determining which graphs are uniquely determined by their spectrum is in general a very hard problem, but all graphs encountered in practice are well behaved and uniquely determined by their dynamics. Hence, our proposed embedding does not loose much information.

For power kernels, it is clear from Theorem~\ref{theo:2} that if two graphs are isomorphic then K(A,B) = 1. Because of the permutation invariance property, we do not have to worry about which ordering of nodes to consider as long as there exist one which gives the required bijection. If two graphs are not isomorphic then their spectrum follow very different behaviors and hence kernel value between them should be much less than 1. To illustrate why our representation satisfies property 2, we use the fact that the spectrum of adjacency matrix is usually very stable under small perturbations, see~\cite{keedwell91}.  Here, the  perturbations means operations like adding or deleting few nodes and edges. It is different from the usual small normed perturbations.  Moreover, our kernel relies on stable statistics such as covariance $\Sigma$ and mean $\mu$  of $\mathbb{S}^A$, which  do not undergo any major jump by small changes in the $\mathbb{S}^A$, assuming the size of graph $n$ is large. Our method thus ensures that small graph perturbations do not lead to any blow up causing relatively big changes in the kernel values.

\begin{figure}[ht]
\begin{center}
\mbox{
\includegraphics[width=3.2in]{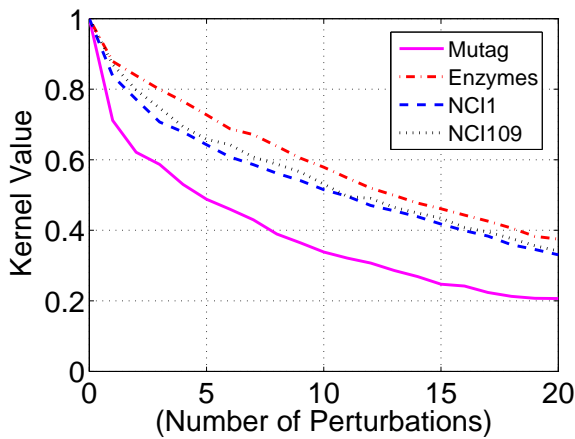}}
\end{center}
\caption{Changes in the value of power kernel with increasing perturbation in the given graph}\label{fig:pert}
\end{figure}

Although, it might be difficult to quantify the sensitivity of power kernels with respect to small perturbations in the graph, we can empirically verify the above claim.  We chose the same four datasets used in the experiments. From each dataset, we randomly sample 100 graphs for the evaluations. We perturb each graph structure by flipping a random edge, i.e., we choose two nodes $i$ and $j$ randomly, if the edge $(i,j)$ was present in the graph then we delete the edge $(i,j)$, otherwise we add the edge $(i,j)$ to the graph. We do this perturbation process 20 times one after the other, thereby obtaining a sequence of 20 graphs with increasing amount of perturbations. After each perturbation we compute the kernel value of the perturbed graph with the original graph. The value of $k$ was again set to be 5. We plot the average kernel values over these 100 points on all the four datasets, in Figure~\ref{fig:pert}.

We can clearly see that the kernel values  smoothly decrease with increasing perturbations.  For MUTAG dataset which consists of smaller graphs, the effect of perturbations is more compared to other datasets with relatively bigger graphs, which is expected. The plots clearly demonstrate that small perturbations do not lead to discontinuous jumps in the kernel values.

\section{Conclusion}

We approached the problem of graph kernels by finding an embedding in functional space. Power kernel, which is based on kernel between these functionals, significantly outperforms the existing state-of-the-art kernels on benchmark graph classification datasets. Our kernel only requires $O(E)$ time to compute and thus this scheme is very practical.

Our focus was to demonstrate the power of an expressive functional representation in a simplest possible way.  We believe that there is a huge scope of improvement in the proposed kernel owing to the possible flexibility in our approach. For example, the choice of Gaussian functions was the natural one. There is a whole room for deriving more expressive functionals on row vectors of $\mathbb{S}^A$, like using kernel density estimators, etc.  Incorporating node and edge label information in power kernel is another area to explore. The idea of discounting subsequent columns of the power iteration could be a useful extension.

We have demonstrated that our functional representation can provide an easy interface for dealing with graphs, a combinatorially hard object. Although we have seen significant gains in the performance over the existing state-of-the-art kernels, in light of the possible future work, we believe a lot more is yet to come.

\section*{Acknowledgement}
The work was partially  supported by  NSF (DMS0808864, SES1131848, III1249316) and  AFOSR (FA9550-13-1-0137).


\begin{thebibliography}{10}

\bibitem{Borgwardt05}
K.~M. Borgwardt and H.~P. Kriegel.
\newblock Shortest-path kernels on graphs.
\newblock In {\em Proceedings of the Fifth IEEE International Conference on
  Data Mining (ICDM 2005)}, pages 74--81, 2005.

\bibitem{Borgwardt205}
K.~M. Borgwardt, C.~S. Ong, S.~Sch{\"o}nauer, S.~V.~N. Vishwanathan, A.~J.
  Smola, and H.~Kriegel.
\newblock Protein function prediction via graph kernels.
\newblock In {\em ISMB (Supplement of Bioinformatics)}, pages 47--56, 2005.

\bibitem{libsvm}
C.-C. Chang and C.-J. Lin.
\newblock Libsvm: A library for support vector machines.
\newblock {\em ACM TIST}, 2(3):27, 2011.

\bibitem{Costa10}
F.~Costa and K.~D. Grave.
\newblock Fast neighborhood subgraph pairwise distance kernel.
\newblock In {\em ICML}, pages 255--262, 2010.

\bibitem{debnath91}
A.~K. Debnath, R.~L.~Lopez de~Compadre, G.~Debnath, A.~J. Shusterman, and
  C.~Hansch.
\newblock Structureactivity relationship of mutagenic aromatic and
  heteroaromatic nitro compounds. correlation with molecular orbital energies
  and hydrophobicity.
\newblock {\em J Med Chem}, 34:786--797, 1991.

\bibitem{farkas2001spectra}
Illes~J Farkas, Imre Der{\'e}nyi, Albert-L{\'a}szl{\'o} Barab{\'a}si, and Tamas
  Vicsek.
\newblock Spectra of “real-world” graphs: Beyond the semicircle law.
\newblock {\em Physical Review E}, 64(2):026704, 2001.

\bibitem{Gartner03}
T.~Gartner, P.~A. Flach, and S.~Wrobel.
\newblock On graph kernels: Hardness results and efficient alternatives.
\newblock In {\em COLT}, pages 129--143, 2003.

\bibitem{jebara10}
T.~Jebara, R.~I. Kondor, and A.~Howard.
\newblock Probability product kernels.
\newblock {\em Journal of Machine Learning Research}, 5:819--844, 2004.

\bibitem{Kashima03}
H.~Kashima, K.~Tsuda, and A.~Inokuchi.
\newblock Marginalized kernels between labeled graphs.
\newblock In {\em ICML}, pages 321--328, 2003.

\bibitem{keedwell91}
A.~D. Keedwell, editor.
\newblock {\em Surveys in Combinatorics}.
\newblock Cambridge University Press, 1991.

\bibitem{Kleinberg98}
J.~M. Kleinberg.
\newblock Authoritative sources in a hyperlinked environment.
\newblock In {\em SODA}, pages 668--677, 1998.

\bibitem{Kondor08}
R.~I. Kondor and K.~M. Borgwardt.
\newblock The skew spectrum of graphs.
\newblock In {\em Proceedings of the 25th International Conference on Machine
  Learning (ICML 2008)}, pages 496--503, 2008.

\bibitem{Kondor02}
R.~I. Kondor and J.~D. Lafferty.
\newblock Diffusion kernels on graphs and other discrete input spaces.
\newblock In {\em ICML}, pages 315--322, 2002.

\bibitem{Kondor09}
R.~I. Kondor, N.~Shervashidze, and K.~M. Borgwardt.
\newblock The graphlet spectrum.
\newblock In {\em Proceedings of the 26th International Conference on Machine
  Learning (ICML 2009)}, pages 529--536, 2009.

\bibitem{kondor2003kernel}
Risi Kondor and Tony Jebara.
\newblock A kernel between sets of vectors.

\bibitem{Kriege12}
N.~Kriege and P.~Mutzel.
\newblock Subgraph matching kernels for attributed graphs.
\newblock In {\em ICML}, 2012.

\bibitem{lin2010power}
Frank Lin and William~W Cohen.
\newblock Power iteration clustering.
\newblock Citeseer.

\bibitem{Mahe04}
P.~Mah{\'e}, N.~Ueda, T.~Akutsu, J.~Perret, and J.~Vert.
\newblock Extensions of marginalized graph kernels.
\newblock In {\em ICML}, 2004.

\bibitem{Schomburg04}
I.~Schomburg, A.~Chang, C.~Ebeling, M.~Gremse, C.~Heldt, G.~Huhn, and
  D.~Schomburg.
\newblock Brenda, the enzyme database: updates and major new developments.
\newblock {\em Nucleic Acids Research}, 32(Database-Issue):431--433, 2004.

\bibitem{Shawe-taylor93}
J.~Shawe-taylor.
\newblock Symmetries and discriminability in feedforward network architectures.
\newblock {\em IEEE Trans. on Neural Networks}, 4:816--826, 1993.

\bibitem{Shervashidze09}
N.~Shervashidze, S.~V.~N. Vishwanathan, T.~Petri, K.~Mehlhorn, and K.~M.
  Borgwardt.
\newblock Efficient graphlet kernels for large graph comparison.
\newblock In {\em Proceedings of International Conference on Artificial
  Intelligence and Statistics. (AISTATS 2009)}, pages 129--143, 2009.

\bibitem{Report:Graph2014}
Anshumali Shrivastava and Ping Li.
\newblock A new space for comparing graphs.
\newblock Technical report, arXiv:1404.4644, 2014.

\bibitem{UganderBK13}
Johan Ugander, Lars Backstrom, and Jon~M. Kleinberg.
\newblock Subgraph frequencies: mapping the empirical and extremal geography of
  large graph collections.
\newblock In {\em WWW}, pages 1307--1318, 2013.

\bibitem{vert2008optimal}
Jean-Philippe Vert.
\newblock The optimal assignment kernel is not positive definite.
\newblock {\em arXiv preprint arXiv:0801.4061}, 2008.

\bibitem{Vishwanathan06}
S.~V.~N. Vishwanathan, K.~M. Borgwardt, and N.~N. Schraudolph.
\newblock Fast computation of graph kernels.
\newblock In {\em NIPS}, pages 1449--1456, 2006.

\bibitem{Vishwanathanjlmr10}
S.~V.~N. Vishwanathan, N.~N. Schraudolph, R.~I. Kondor, and K.~M. Borgwardt.
\newblock Graph kernels.
\newblock {\em Journal of Machine Learning Research}, 11:1201--1242, 2010.

\bibitem{vishwanathanSV07}
S.~V.~N. Vishwanathan, A.~J. Smola, and R.~Vidal.
\newblock Binet-cauchy kernels on dynamical systems and its application to the
  analysis of dynamic scenes.
\newblock {\em International Journal of Computer Vision}, 73(1):95--119, 2007.

\bibitem{Wale06}
N.~Wale and G.~Karypis.
\newblock Comparison of descriptor spaces for chemical compound retrieval and
  classification.
\newblock In {\em ICDM}, pages 678--689, 2006.

\bibitem{wolf03}
Lior Wolf and Amnon Shashua.
\newblock Learning over sets using kernel principal angles.
\newblock {\em The Journal of Machine Learning Research}, 4:913--931, 2003.

\end{thebibliography}

\end{document}